\documentclass[letterpaper, 10 pt, conference]{ieeeconf}  

\IEEEoverridecommandlockouts                              

\usepackage{amsmath} 
\usepackage{amssymb}  
\usepackage{hyperref}
\usepackage{algorithm}
\usepackage{graphicx}
\usepackage[noend]{algpseudocode}

\usepackage{colortbl} 
\usepackage{booktabs}
\title{\LARGE \bf
Manipulate-to-Navigate: Reinforcement Learning with Visual Affordances and Manipulability Priors
}

\author{Yuying Zhang$^{1}$ and Joni Pajarinen$^{1}$
\thanks{*This work was supported by the Chinese Scholarship Council and by the European Union's Horizon Europe Project XSCAVE under grant agreement No 101189836}
\thanks{$^{1}$ Yuying Zhang and Joni Pajarinen are with the Aalto Robot Learning lab, Department of Electrical Engineering and Automation, Aalto University, Espoo, Finland
        {\tt\small \{yuying.zhang, joni.pajarinen\}@aalto.fi}}%
}

\begin{document}

\maketitle
\thispagestyle{empty}
\pagestyle{empty}

\begin{abstract}
Mobile manipulation in dynamic environments is challenging due to movable obstacles blocking the robot’s path. Traditional methods, which treat navigation and manipulation as separate tasks, often fail in such "manipulate-to-navigate" scenarios, as obstacles must be removed before navigation. In these cases, active interaction with the environment is required to clear obstacles while ensuring sufficient space for movement.
To address the manipulate-to-navigate problem, we propose a reinforcement learning-based approach for learning manipulation actions that facilitate subsequent navigation. Our method combines manipulability priors to focus the robot on high manipulability body positions with
affordance maps for selecting high-quality manipulation actions. By focusing on feasible and meaningful actions, our approach reduces unnecessary exploration and allows the robot to learn manipulation strategies more effectively.

We present two new manipulate-to-navigate simulation tasks called Reach and Door with the Boston Dynamics Spot robot. The first task tests whether the robot can select a good hand position in the target area such that the robot base can move effectively forward while keeping the end effector position fixed. The second task requires the robot to move a door aside in order to clear the navigation path. Both of these tasks need first manipulation and then navigating the base forward. Results show that our method allows a robot to effectively interact with and traverse dynamic environments. Finally, we transfer the learned policy to a real Boston Dynamics Spot robot, which successfully performs the Reach task.
\end{abstract}

\section{Introduction}

Mobile manipulators combine advanced mobility with dexterity, making them ideal candidates for a wide range of applications such as human assistance, manufacturing, and agriculture. These systems are capable of performing complex tasks that involve both navigation and manipulation. However, a significant challenge arises from the complexity of whole-body control~\cite{wholebody_2022,spot_goandfetch_2021}, particularly for legged robots that have a high number of degrees of freedom. This complexity makes it difficult to effectively coordinate the movements of the robot's body and arms in dynamic environments.

\begin{figure}[thb]
    \centering
    \includegraphics[width=0.48\textwidth, height=0.16\textwidth]{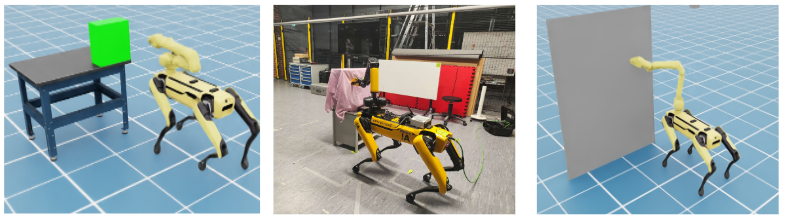}
    \caption{\textbf{Illustration of manipulate-to-navigate tasks.} \textit{Left: Spot-Reach-Sim:} The robot selects a target end-effector within the green target area. Once positioned, the robot advances its base forward while maintaining the end-effector's position. \textit{Middle: Spot-Reach-Real:} We set up a similar reach task in the real world with the Boston Dynamics Spot robot. In this task, the robot needs to position the end-effector in the purple area on the table first and then navigate forward with the given velocity. \textit{Right: Spot-Door-Sim:} The sliding gray door moves along the y-axis, while the robot moves forward along the x-axis. The door has a spring that tries to close the door. The robot must find an arm pose to push the door aside and then pass through while keeping its hand in place to prevent the door from returning back.
    }
\label{fig:task}
\end{figure}

Most existing approaches treat body control and arms control as distinct tasks, allowing for the integration of existing advancements in both manipulation and navigation~\cite{xiali2020relmogen,  NavManiCosts2022}. However, this separation fails to capture the critical collaboration required between these two processes. The lack of this separation becomes particularly obvious in dynamic environments where obstacles, such as doors, curtains, or wheeled chairs, tend to return to their original positions after being manipulated. These scenarios demand interactive navigation, where the robot must dynamically manipulate obstacles while ensuring smooth navigation. Therefore, an integrated approach that fosters collaboration between manipulation and navigation is essential for effective mobile manipulation in environments with dynamic obstacles.

While traditional methods often focus on planning and coordination~\cite{adaskill_mm} or use inverse reachability maps~\cite{2009reachable_arm, NavManiCosts2022} to guide the robot’s movements, these techniques are typically limited in their ability to adapt to dynamic, real-world environments. Although inverse reachability maps offer a way to predict areas the robot can reach based on its kinematic structure, they lack the generalizability needed in complex settings. Reinforcement learning (RL), which has demonstrated promise in generalizing to dynamic environments, is typically trained in an end-to-end manner. However, this requires extensive exploration, which is time-consuming and impractical for real-world robots. Furthermore, RL methods are often conducted in simplified environments, making it challenging to transfer learned policies to more complex robotic structures operating in real-world scenarios.

To address these challenges, this paper proposes the integration of manipulability priors and visual affordance maps to improve the sample efficiency and safety of exploration in RL. Manipulability priors, based on the robot's kinematic structure, ensure the feasibility of the robot's movements, while affordance maps guide the robot’s interaction with objects and obstacles, accelerating the learning process. The combination of these priors with RL enhances exploration safety and facilitates the transfer of RL to more complex, real-world robotic systems.
The key contributions of this paper are as follows:
\begin{itemize}
    \item We formulate and address the novel manipulate-to-navigate task, where a robot must actively manipulate obstacles to enable subsequent navigation.
    \item We propose a reinforcement learning framework that integrates manipulability priors and training-free visual affordance maps, which together reduce unnecessary exploration and guide the robot toward feasible and effective manipulation actions.
    \item We design two new manipulate-to-navigate benchmark tasks, named Reach and Door, in a simulation environment with the Boston Dynamics Spot robot, capturing the essential challenges of interactive navigation.
    \item We demonstrate successful sim-to-real policy transfer, showing that a policy trained in simulation can generalize to a real-world Reach task using the physical Spot robot.
\end{itemize}

\section{Related Work}
Mobile manipulation systems face ongoing challenges in interactive navigation due to environmental uncertainties, the high degree of freedom, and the need for optimal coordination ~\cite{surveyleggedmm,surveywheelmm}. Classical approaches have leveraged known system information ~\cite{2009reachable_arm, 2012manipulability, 2024manicapability} and formulated mobile manipulation as a motion and path planning problem ~\cite{NavManiCosts2022}. However, these methods primarily focus on well-structured environments and often suffer from poor generalization to unseen scenarios.

To improve the generalization, several learning-based methods have been proposed to optimize base pose selection for grasping, aiming to maximize reachability or minimize task time by reducing unnecessary repositioning~\cite{reaprior_2022Jan, NavManiCosts2022, pregrasp_2024}, but none explicitly optimize the end-effector pose for base movement. While Reinforcement Learning (RL) enables robots to solve complex tasks in dynamic environments, its high training cost due to the expensive exploration process remains a major challenge. To address this, \cite{reaprior_2022Jan} uses a reachability map as behavior priors, which accelerate training and guide exploration. The Kullback-Leibler (KL) divergence is a common technique to incorporate priors, which restricts exploration from deviating excessively from the prior distribution~\cite{KLloss_2008}. Building on this idea, we propose using a manipulability map, derived from the robot’s configuration, as a behavior prior for efficient arm pose generation. This map guides exploration by constraining the action space, improving sample efficiency, and facilitating faster convergence.

Affordance defines the possible actions an environment affords to an agent ~\cite{affsurvey_rl_2020}, providing an action-oriented characterization of the agent’s state ~\cite{aff_gibson}. Recent works have leveraged affordance models to learn high-quality state-action spaces, accelerating skill acquisition in tasks such as dexterous grasping ~\cite{affgrasp_2021mandikal, affDQN_2019}, navigation ~\cite{baseplacement,mm_base_feasibility}, and static manipulation ~\cite{affvlm_2024, affrl_mani2023}. Affordance-based methods can also reduce the state-action space, leading to more efficient and precise learning. Several algorithms have focused on learning affordance models from play data ~\cite{affpred_2023belkhale, modelaff_rl2022} or incorporating multi-view observations to predict action effects ~\cite{farmani_aff_2024}. These affordance models have been used to train goal-conditioned policies with visual input ~\cite{visual_aff_il2021} or integrate language instructions for solving complex manipulation tasks ~\cite{affvlm_2024, Retrieval_aff_2024}. However, most of these methods require additional training to learn the affordance model. Instead, we leverage a visual foundation model to segment affordance maps directly. This map provides pixel-level manipulation candidates, guiding RL exploration while reducing the action space and accelerating training.

In summary, this paper focuses on a novel task, manipulate-to-navigate, selecting the end-effector position with the consideration of base reachability. We address this task by integrating visual affordance and manipulability into an RL framework, which significantly reduces training time while maintaining high task performance.

\begin{figure}[tb]
    \centering
    \includegraphics[width=0.48\textwidth, height=0.20\textwidth]{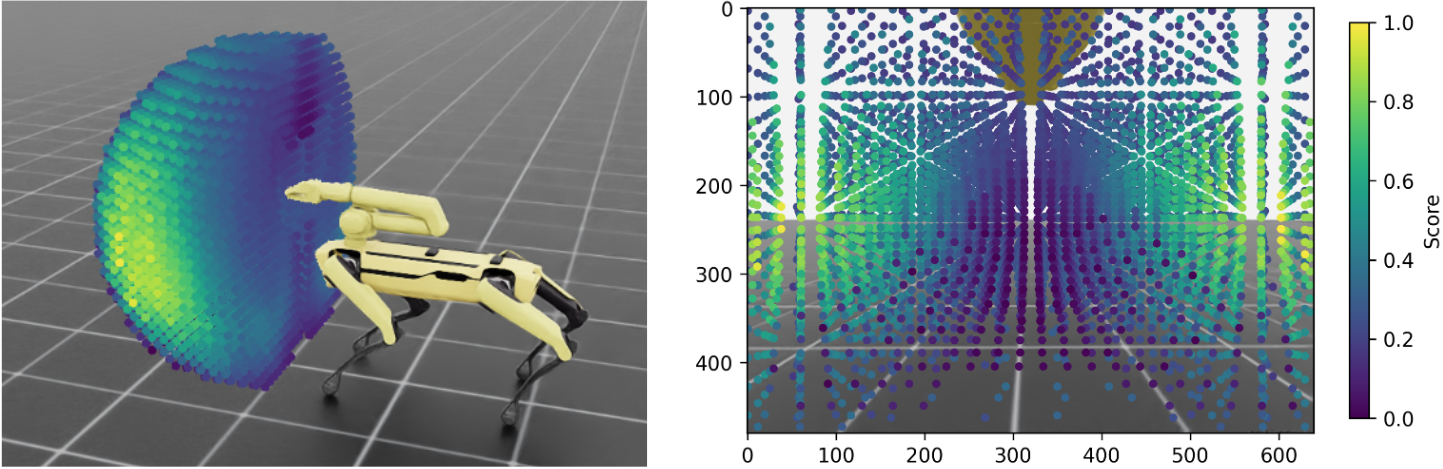}
    \caption{\textbf{Visualization of a manipulability map for the Spot robot.} On the left, we show the normalized manipulability score for each hand position in 3D space with different colors. On the right, we project the end-effector position into the image space of the camera mounted on the robot body. A higher score suggests greater dexterity and freedom of movement in the current configuration, while a smaller value indicates proximity to singularities, where control and movement become more constrained.}
\label{fig:visual_map}
\end{figure}

\section{Preliminaries}
\subsection{Problem Statement}

Given the image from its onboard camera, our objective is to determine the end-effector position that both satisfies the task requirements and maximizes the base’s flexibility. To efficiently discover such poses, we incorporate behavior priors that consider the robot's structure and workspace, specifically its reachability and manipulability. Additionally, we construct an affordance map based on the visual input to improve sample efficiency and accelerate the reinforcement learning (RL) process. While most affordance models require training, particularly when using visual input, we leverage a visual foundation model to extract information from objects without additional training.

\subsection{RL with action prior}

The Markov Decision Process (MDP) is defined by the tuple \( (S, A, P, R, \gamma, P_0 ) \), where \( S \) and  \( A \) are the state space and the action space, respectively. \( P: S \times A \times S \to \mathbb{R}^{+} \) is the transition probability function, where \( P(s_{t+1})|s_t,a_t) \) denotes the probability of transitioning from state \( s_t \) to state \( s_{t+1} \) after taking action \( a_t \), \( R(s_t,a_t) \) is the reward function, providing the expected reward for executing action \( a_t \) in state \( s_t \), \( \gamma \in (0,1] \) is the discount factor, controlling the weight of future rewards. \(  P_0(s) \) is the initial state distribution.

The goal of reinforcement learning (RL) is to find an optimal policy \( \pi^* \)that maximizes the expected cumulative reward. The Q-value function for a given policy \( \pi \) is defined as:

\begin{align}
Q^\pi(s) = \mathbb{E_{ \pi}} \left[ \sum_{t=0}^{\infty} \gamma^t R(s_t,a_t) \Bigg| s_t = s , a_t=a\right]
\end{align}

where \( \mathbb{E} \) represents the expected value, \( s_t \) is the state at time \( t \), and \( a_t \) is the action taken at time \( t \) under policy \( \pi \). To find the optimal policy \( \pi^* \), we maximize the Q-value:

\begin{align}
\pi^* = \arg \max_\pi Q^\pi(s)
\end{align}

Given an action prior distribution \( p(a_t|c_r) \), where  \( c_r \) is the robot configuration, the learning objective turns into a relative-entropy policy optimization problem, the KL regularization between the agent policy and prior policy is taken to achieve this goal, hence, the loss function can be noted as:

\begin{equation} 
\mathcal{L}(\theta) = \mathbb{E} \left[ \left( Q_{\theta}(s, a) - y \right)^2 \right] + \lambda \sum_{t=0}^{\infty} \gamma^t \text{KL} \left[ \pi(a_t | s_t) | p(a_t | c_r) \right] 
\label{eq:loss}
\end{equation}
\begin{equation}
    y = R(s, a) + \gamma Q_{\theta'}(s_{t+1}, \arg\max_{a_{t+1}} Q_{\theta}(s_{t+1}, a_{t+1}))
    \label{eq:target}
\end{equation}
where \( Q_{\theta}(s, a) \) is the current Q-value estimate from the neural network.\( \lambda \) is a weighting factor for KL regularization.

\subsection{Affordance}

Affordance refers to the potential actions or behaviors that an object, environment, or surface affords to an agent based on its properties and the agent's capabilities. In the context of robotic manipulation, affordance information can be extracted from visual input by leveraging a visual foundation model. This affordance information can then be used to calculate the affordance Q-value by combining the affordance data derived from the visual input with the Q-value function: 
\begin{equation} 
s^{\text{aff}} = \Phi(I_{\text{rgbd}})
\end{equation}
\begin{equation} 
 s = \psi(I_{\text{rgbd}}) 
\end{equation}
Therefore, the Q function is represented as \(Q(s, a,s^{\text{aff}}; \theta) \)

\begin{figure*}[tb]
    \centering
    \includegraphics[width=0.9\textwidth, height=0.38\textwidth]{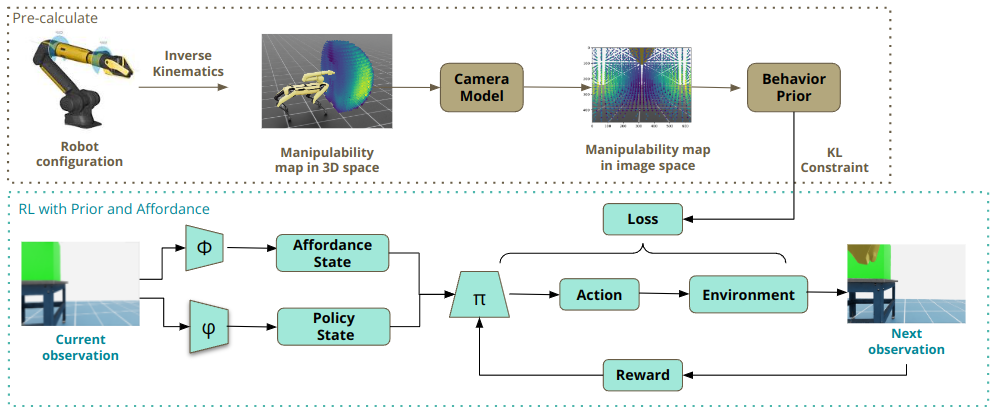}
    \caption{\textbf{Pipeline for our method.} 
    The upper section illustrates the pre-calculation phase, where we generate manipulability maps in 3D space according to the robot configuration, filtering out positions that result in inverse kinematic failures or fall outside the camera's field of view. This 3D map is then projected into pixel space using the camera model (comprising intrinsic and extrinsic matrices) to create a manipulability map in image space. This map is used to calculate the final behavior prior, which is incorporated into the loss function alongside a KL constraint during the training phase. The lower section depicts the reinforcement learning process, which integrates visual affordances derived from the current observation. Separate encoders (\(\phi\) and \(\psi\)) extract policy states \(s_t\) and affordance states \(s^\text{aff}\), respectively, which are input to the policy to generate optimized actions that maximize environmental rewards. This two-stage approach enables the robot to leverage spatial manipulability knowledge while learning effective interaction strategies.
    }
\label{fig:framework}
\end{figure*}

\section{Method}

\begin{algorithm}[t]
\caption{Our Algorithm}
\label{alg:ours}
\begin{algorithmic}[1]
\State \textbf{Initialize:} 
\State \quad \( p(a_t | c_r)\) \quad Calculate Manipulability prior distribution with the robot configuration
\State \quad \(\theta, \theta' \gets \theta_0\) \quad Initialize current network parameters
\State \quad \( D \gets \emptyset \) \quad \text{Initialize replay memory} 
\State \quad \(\epsilon \gets \epsilon_0\) \quad \text{Initialize exploration rate} 

\State  \textbf{Repeat} for each episode:
\State  \quad Initialize state \( s_0 \) \\
\quad \textbf{For} each timestep \( t \):
\begin{itemize}
    \item \( s_t^{\text{aff}} = \Phi(\mathbf{I_t}) , s_t = \psi(\mathbf{I_t}) \) where \( I_t \) is the RGB-D input image
    \item \( a_t = \arg\max_a Q(s_t, a,s_t^{\text{aff}}; \theta) \) with \(\epsilon\)-greedy policy
    \item Take action \( a_t \), observe reward \( r_t \) and next state \( s_{t+1} \)
    \item Store transition \( (s_t,s_t^{\text{aff}}, a_t, r_t, s_{t+1}, s_{t+1}^{\text{aff}} \) in replay memory \( D \)
    \item  \( B = \{(s_i,s_i^{\text{aff}}, a_i, r_i, s_{i+1},  s_{i+1}^{\text{aff}}\} \) from replay memory
    \item Compute target value according to Eq.~\ref{eq:target}

    \item Perform gradient descent step on loss function Eq.~\ref{eq:loss}
    \item Update the parameters \( \theta \) of the current Q-network
    \item \( \theta^- \leftarrow \theta, \epsilon^- \leftarrow \epsilon \)
\end{itemize}
\textbf{Until} convergence \\
\Return \( \pi^* = \arg \max_a Q(s, a,s^{\text{aff}}; \theta) \)
\end{algorithmic}

\end{algorithm}
Every robot has inherent hardware limitations, such as structural constraints and degrees of freedom. Inspired by previous studies~\cite{2009reachable_arm, 2012manipulability,reaprior_2022Jan}, we pre-compute the robot’s operational workspace using structural information as an action prior to guiding the training of the agent. This manipulability map is generated offline, considering joint limitations and self-collisions in 6D space, ensuring that selected actions remain feasible. In detail, we first calculate all the possible end-effector positions and the manipulability of each position according to the hardware configuration of the spot robot. The manipulability quantifies a robot manipulator's ability to move and exert forces in various directions from a given configuration which provides a measure of how effectively a robot can perform tasks involving positioning and orienting its end-effector. It is defined as:

\begin{equation}
    w = \sqrt{\det(J J^T)}
\end{equation}

where \(J\) is the Jacobian matrix of the robotic manipulator. After calculating the potential hand positions of the robot and its manipulability for each position, we first filter out the candidate positions that lead to inverse kinematic (IK) failures to ensure only feasible configurations are retained. The remaining valid end-effector positions are then projected into the image (pixel) space of a forward-facing camera mounted on the robot.

This projection is performed using both the intrinsic and extrinsic parameters of the camera. Specifically, we use the extrinsic matrix \( [R \mid t] \), where \( R \in \mathbb{R}^{3 \times 3} \) is the rotation matrix and \( t \in \mathbb{R}^{3} \) is the translation vector, to transform the 3D world coordinates of the candidate hand positions into the camera coordinate frame. Then, we apply the intrinsic matrix \( K \in \mathbb{R}^{3 \times 3} \), which includes the focal lengths and principal point of the camera, to project the camera-frame coordinates onto the 2D image plane:

\begin{equation}
\begin{bmatrix}
u \\
v \\
1
\end{bmatrix}
=
K \cdot
\left[ R \mid t \right] \cdot
\begin{bmatrix}
x \\
y \\
z \\
1
\end{bmatrix}
\end{equation}

Here, \( (x, y, z) \) are the 3D world coordinates of a candidate hand position, and \( (u, v) \) are the resulting pixel coordinates in the image space. Only those 3D positions that successfully map within the image bounds are considered as valid action candidates in the pixel-based action space. Fig.~\ref{fig:visual_map} visualize the normalized score in different colors. The yellow points indicated higher manipulation capability, which means more dexterity and freedom of movement in the current configuration, while the point in blue color suggests having lower dexterity and the movement is more constrained. We assume that the higher manipulability also indicates higher flexibility for the base movement when maintaining the hand position. Therefore, we use the manipulability of the hand position as the behavior prior to guiding the training of the agent. Furthermore, we employ the KL constraint between the current agent policy and the prior policy.   

For solving tasks involving dynamic movable obstacles, we additionally extract affordance features from visual input to enhance the sample efficiency of exploration. We use the image \( I_\text{rgbd} \in \mathbb{R}^{480 \times 640 \times 4} \) from the front camera mounted on the robot as the observation, where the image size is \( 480 \times 640 \) and each pixel contains four channels (RGB and depth). The policy state \(s_t  \in \mathbb{R}^{348}\) is extracted by the advanced visual foundation model Dinov2~\cite{dinov2} and the task-related affordance state \(s_t^\text{aff}  \in \{0,1\}^{480 \times 640}\)is segmented through Mobile-Segment Anything(SAM)~\cite{mobile_sam}. Unlike traditional affordance models that predict action outcomes based on training data, our approach directly leverages the concept of affordance—defining whether an object can be manipulated rather than predicting the effect of an action. The affordance map is computed online based on the training-free models, dynamically adapting to the task and environment.

To combine the precomputed manipulability map (derived from the robot's configuration) with online affordance information (extracted from visual input) smoothly, we select the pixel space as the action space. Therefore, the action at time step  \( t \), denoted as \( a_t \) is defined as the pixel coordinates:\({a_t} = (I_x, I_y) \in \{0, \dots, 639\} \times \{0, \dots, 479\}\). The policy generates the optimal hand position in image coordinates, and the generated action is then projected to the 3D world coordinates using the known camera parameters, which can be executed by the robot.

Our proposed framework is compatible with both discrete and continuous reinforcement learning algorithms. In this work, we design a task-specific reward function to guide learning, defined as:
\begin{equation} 
\begin{split}
r(s,a) &= w_1r_\text{ik}(s, a)+w_2r_\text{balance}(s, a) + \\ 
&w_3r_\text{arm}(s, a)+w_4(r_\text{move}(s, a)*reach)
\end{split}
\end{equation}

where \(w_1,w_2,w_3,w_4\) are weights scaling the rewards. 

\begin{itemize}
  \item \( r_\text{ik}(s, a) \) penalizes inverse kinematics failure, ensuring that the selected end-effector positions are kinematically feasible.
  \item \( r_\text{balance}(s, a) \) penalizes base instability, encouraging safer configurations.
  \item \( r_\text{arm}(s, a) \) is task-specific rewards for the arm pose.
  \item \( r_\text{move}(s, a) = \text{Dist}(s, a) \) measures the translational distance the base can travel while maintaining a fixed arm pose, used only when the hand has reached the target (i.e., \(\text{reach} = 1\)). Note that the rotational distance is not considered here.
\end{itemize}

In summary, our method learns an optimal end-effector position in image space that maximizes the base's mobility while accounting for interaction constraints. By combining manipulability priors and affordance maps, we constrain the action space to promote feasible and effective interactions. This improves sample efficiency and ensures that selected actions are both reachable and meaningful for the robot. Algorithm ~\ref{alg:ours} shows the details of the algorithm.

\section{Experiments}
To evaluate our proposed method, we design two manipulate-to-navigate tasks in the Isaac Sim simulator using the Boston Dynamics Spot robot. These tasks are created to assess the effectiveness of integrating manipulability priors and affordance maps in reinforcement learning. We conduct a comprehensive ablation study to analyze the contribution of each component in our framework. The evaluation is based on two key metrics: success rate (whether the robot completes the task) and base movement distance (the distance the robot base can advance after manipulation). These metrics reflect both task completion and how well the robot prepares the environment for navigation through effective manipulation.

\begin{figure}[tb]
    \centering
    \includegraphics[width=0.45\textwidth, height=0.3\textwidth]{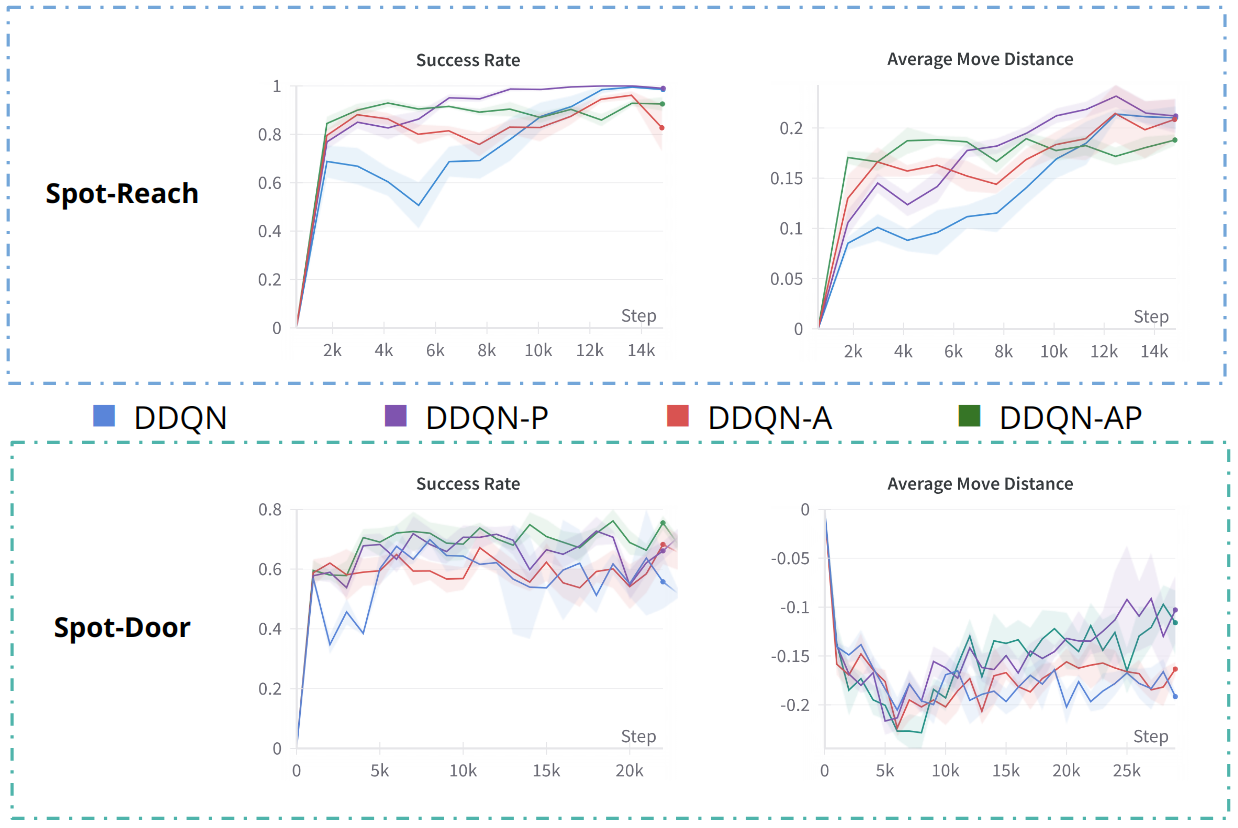}
    \caption{\textbf{Success rate and average base movement distance curves of our methods and baseline.} 
   The first column presents the training results for the Reach task, while the second column shows results for the Door task. For each task, the left plot compares different approaches, including DDQN, DDQN-P, DDQN-A, and DDQN-AP, in terms of success rate over training steps, whereas the right plot compares the approaches based on the average base movement distance. The results are averaged over five random seeds, with the shaded regions representing the standard error. 
    }

\label{fig:suc_rate}
\end{figure}

\subsection{Tasks}

\textbf{Spot-Reach}: In this task, there is a target area on the table shown in green color, and the table is on the left side of the robot. In this task, the robot first selects an optimal pose to position its hand within the green target area. Once the robot hand reaches the target area, the robot base moves forward and passes the table while keeping the end effector in the selected position. The task is shown on the left of Fig.~\ref{fig:task}. We use the reward of base move distance to encourage the selection of a more flexible hand position. The task is defined as solved as long as the hand reaches the target area in green. Note that we only consider the width and height into count without the depth of the rectangle.
 
\textbf{Spot-Door}: As shown in the middle of Fig.~\ref{fig:task}, a gray door with a prismatic joint along the y-axis is positioned in front of the robot. Consequently, the door can only move sideways and tends to return to its original position, similar to a real-world pull door. To pass through, the robot must first push the door aside and then maintain its hand position to ensure sufficient space for navigation.

To evaluate task completion, we use a visual feature-based metric. Specifically, the task is considered solved when the visible area of the door in the camera view falls below a predefined threshold: \(L_\text{door} < L_\text{threshold} \). \(L_\text{door}\) represents the length of the door in the image, computed via segmentation from the visual input using a pre-trained model~\cite{mobile_sam}, while \(L_\text{threshold}\) is a hyperparameter defining the threshold.

\textbf{Spot-Reach-Real}: As shown in Fig.~\ref{fig:task} on the right, we set up the Reach task in the real world to evaluate the transferability from simulation to reality. The purple cloth on the table serves as the target for the robot’s hand. Initially, the robot is positioned in front of the table, with the target located to the left of the robot. The task is considered solved if the robot successfully grasps the cloth. Notably, the grasp action is automated as long as the robot’s hand is in proximity to the target object. Upon successfully grasping the cloth, forward velocity is applied to assess the reachability of the robot’s base. To facilitate sim-to-real transfer, we use an identical camera model in both simulation and the real world. The only necessary adjustment is physically mounting the camera on the real robot in the same position and orientation as in the simulation setup, specifically in front of the robot.

\subsection{Comparison}

Our proposed framework is compatible with both discrete and continuous reinforcement learning algorithms. In this work, we use Double Deep Q-Network (DDQN)~\cite{DDQN} as a representative example due to its training stability, ease of implementation, and lower sample complexity. To evaluate the efficiency of manipulability prior and the visual affordance separately for manipulate-to-navigate tasks, we implement our methods based on DDQN and compare against the following: \textit{i.} Pure DDQN (DDQN); \textit{ii.} DDQN with manipulability prior only(DDQN-P); \textit{iii.} DDQN with visual affordance only(DDQN-A); \textit{iv.} our algorithm, DDQN with both manipulability prior and visual affordance(DDQN-AP). We train all agents with 4 parallel environments using a single RTX 3080 graphics card.

\begin{table}[tb]
\centering
\caption{Success rate of the methods in real world with 10 trails.}
\begin{tabular}{*5c}\toprule
\textbf{Algorithm} & \textbf{DDQN} &  \textbf{DDQN-P} & \textbf{DDQN-A} & \textbf{DDQN-AP}   \\ \midrule
\rowcolor{white}
    \textbf{Success Rate}        &      0.1  &      0.1     &         0.8          &0.8 \\   
\bottomrule
\end{tabular}
\label{tab:sr}

\end{table}

Fig.~\ref{fig:suc_rate} shows the comparison result based on these two novel tasks. The first column presents the training results for the Reach task, while the second column shows results for the Door task. For each task, the left plot compares different approaches, including DDQN, DDQN-P, DDQN-A, and DDQN-AP, in terms of success rate over training time steps, whereas the right plot compares the approaches based on the average base movement distance. Additionally, only translational distance is considered, excluding rotational changes. The results are averaged over five random seeds, with the shaded regions representing the standard error.
As shown, DDQN exhibits slower convergence and higher variance. In contrast, DDQN-A, DDQN-AP, and DDQN-P achieve the highest overall performance, maintaining a consistently high success rate throughout training. As expected, DDQN-AP demonstrates the fastest convergence, reaching over 85\% success rate within 1000 training time steps and converging around 4000 time steps for the reach task, while DDQN requires more than three times as many training time steps. These findings suggest that incorporating manipulability priors (P) and visual affordance guidance (A) enhances learning efficiency in mobile manipulation while motivating the robot to select a hand position with higher base reachability(only 0.03 meters different).

For the Door task, methods DDQN-A, DDQN-AP, and DDQN-P also achieve relatively high success rates more efficiently than DDQN, showing the efficiency for solving complex tasks. However,  achieving a positive average move distance remains challenging. This is primarily because the robot needs to move backward to maintain balance while interacting with the sliding door, as the door only moves along the y-axis, but the robot experiences force along the x-axis during manipulation. Despite this, the overall base movement distance increases, indicating that the robot actively selects positions with higher base reachability. Introducing a small penalty for negative movement in the reward function may reduce unnecessary retreating while preserving essential balancing behavior.

Table ~\ref{tab:sr} presents the success rates of different methods across 10 trials in the real world. The baseline DDQN model, which lacks any prior guidance, achieves a low success rate of 0.1. As expected, methods that incorporate an affordance map demonstrate better transferability from simulation to reality. With only camera position adaptation, both the DDQN-A and DDQN-AP methods achieve 8 successful trials out of 10. One failure occurred due to a perception error, where the robot mistakenly grasped the table beneath the target. Additionally, among the successful trials, the agent trained with the DDQN-AP method achieved a maximum base movement distance of 0.5 meters.

Incorporating either manipulability priors (DDQN-P) or affordance maps (DDQN-A) alone shows divergent results. While DDQN-P still performs poorly with a success rate of 0.1, DDQN-A significantly improves performance to 0.8. This suggests that affordance maps are more informative in guiding the agent toward feasible and meaningful interactions with the environment, especially in the early stages of learning. Most notably, the combined approach (DDQN-AP), which integrates both manipulability priors and affordance maps, maintains the high success rate of 0.8. This demonstrates that manipulability priors can be incorporated without degrading performance, and may help ensure kinematically feasible hand positions during deployment, even if they offer limited advantage alone during evaluation. These results support our hypothesis that visual affordance guidance is critical for effective exploration, while manipulability priors act as a complementary filter to ensure physical feasibility.

Analyzing failure cases for both Reach and Door tasks, we observe that a key source of failure in methods using visual affordance guidance stems from perception errors. Specifically, these are primarily introduced by the segmentation model used to extract affordance cues. Inaccurate or incomplete segmentation can lead to poor action selection. This limitation can potentially be mitigated by fine-tuning the perception module on task-specific data or integrating semantic information to improve robustness and accuracy.

Another major challenge arises from camera viewpoint changes during manipulation. As the robot adjusts its leg pose to maintain balance, particularly when external forces are applied, the camera mounted on the robot also shifts. This issue is especially evident in the Door task, where the force exerted to slide the door often causes the robot to walk backward, triggering significant leg movements to re-stabilize the base. These adjustments alter the camera's viewpoint, leading to misalignment between the perceived scene and the actual environment. As a result, discrepancies arise between the planned and executed end-effector trajectories, frequently leading to manipulation failures. This challenge could potentially be mitigated by incorporating correction mechanisms during execution to adapt to viewpoint changes in real time.

An additional source of failure stems from inaccuracies in the image-to-world projection process. Since actions are selected in the image coordinate space and then projected into the real world, precise camera calibration and transformation models are essential. However, even minor misalignments in intrinsic or extrinsic camera parameters—exacerbated by the robot's dynamic movements—can lead to projection errors. These inaccuracies distort the estimated hand positions, resulting in imprecise or failed manipulation attempts. Moreover, errors in depth estimation further compound the problem by introducing discrepancies between predicted and actual interaction points. To address this, we plan to improve robustness in future work by redefining the action space from image coordinates to world coordinates, thereby reducing dependence on projection accuracy.

\section{Conclusion}

This paper targets a novel task, manipulate-to-navigate, where the robot must first manipulate objects to clear a path for base navigation. Additionally, the tendency of obstacles to return to their original positions increases the complexity of the task. To address this challenge, we propose a method that enables the robot to move obstacles and hold them in place to prevent collisions before navigation. Our approach generates end-effector poses by leveraging affordance maps and manipulability priors. The affordance map, constructed using a training-free visual foundation model, accelerates the learning process and enhances transferability. Meanwhile, the manipulability prior is incorporated via KL constraints to guide the training process effectively. We developed two manipulate-to-navigate tasks, Reach and Door, in the simulation to evaluate our framework. Results show that our approach improves sample efficiency and accelerates RL training, achieving a success rate of over 85\% within approximately 1,000 training time steps for the Reach task. Finally, we validate our method in real-world experiments, achieving an 80\% success rate in the Reach task and 0.5 meter base moving distance at maximum.

For future work, we plan to refine our framework along two key directions. First, we aim to transition the action space from image coordinates to 3D world coordinates, eliminating the dependence on image-to-world projection and improving manipulation precision.  Second, we seek to enhance the generalizability of our approach across a broader range of manipulation and navigation tasks, combining task-conditioned policies.

\addtolength{\textheight}{-12cm}   


\bibliographystyle{IEEEtran}
\bibliography{IEEEabrv}

\end{document}